%% file: main.tex
\documentclass[sigconf]{acmart}

\usepackage{bm}
\usepackage{multirow}
\usepackage{color,soul}
\usepackage{bbm}
\usepackage{array}

\newcommand{\ourModel}{GraphHAM}

\AtBeginDocument{%
  \providecommand\BibTeX{{%
    \normalfont B\kern-0.5em{\scshape i\kern-0.25em b}\kern-0.8em\TeX}}}

\renewcommand\footnotetextcopyrightpermission[1]{} 
\setcopyright{none} 

\begin{document}

\title{Graph Embedding with Hierarchical Attentive Membership}

\author{Lu Lin, Ethan Blaser, Hongning Wang}
\affiliation{
    \department{Department of Computer Science}
    \institution{University of Virginia} 
    \city{Charlottesville}
    \state{VA 22904}
    \country{USA}
 }
\email{{ll5fy, ehb2bf, hw5x} @virginia.edu}


\begin{abstract}
The exploitation of graph structures is the key to effectively learning representations of nodes that preserve useful information in graphs. 
A remarkable property of graph is that a latent hierarchical grouping of nodes exists in a global perspective, where each node manifests its membership to a specific group based on the context composed by its neighboring nodes. 
Most prior works ignore such latent groups and nodes' membership to different groups, not to mention the hierarchy, when modeling the neighborhood structure. Thus, they fall short of delivering a comprehensive understanding of the nodes under different contexts in a graph.

In this paper, we propose a novel hierarchical attentive membership model for graph embedding, where the latent memberships for each node are dynamically discovered based on its neighboring context. Both group-level and individual-level attentions are performed when aggregating neighboring states to generate node embeddings.
We introduce structural constraints to explicitly regularize the inferred memberships of each node, such that a well-defined hierarchical grouping structure is captured.
The proposed model outperformed a set of state-of-the-art graph embedding solutions on node classification and link prediction tasks in a variety of graphs including citation networks and social networks. 
Qualitative evaluations visualize the learned node embeddings along with the inferred memberships, which proved the concept of membership hierarchy and enables explainable embedding learning in graphs.
\end{abstract}



\maketitle

\pagestyle{plain} 

\vspace{-3mm}
\input{intro.tex}
\input{related.tex}
\input{method.tex}

\input{experiment.tex}
\input{conclusion.tex}

\begin{acks}
This paper is based upon the work supported by the National Science Foundation under grant IIS-1553568, IIS-1718216 and IIS-2007492.
\end{acks}

\bibliographystyle{ACM-Reference-Format}
\bibliography{reference}

\end{document}

%% file: intro.tex
\section{Introduction}

\begin{figure} 
\centering
\includegraphics[width=3.5in]{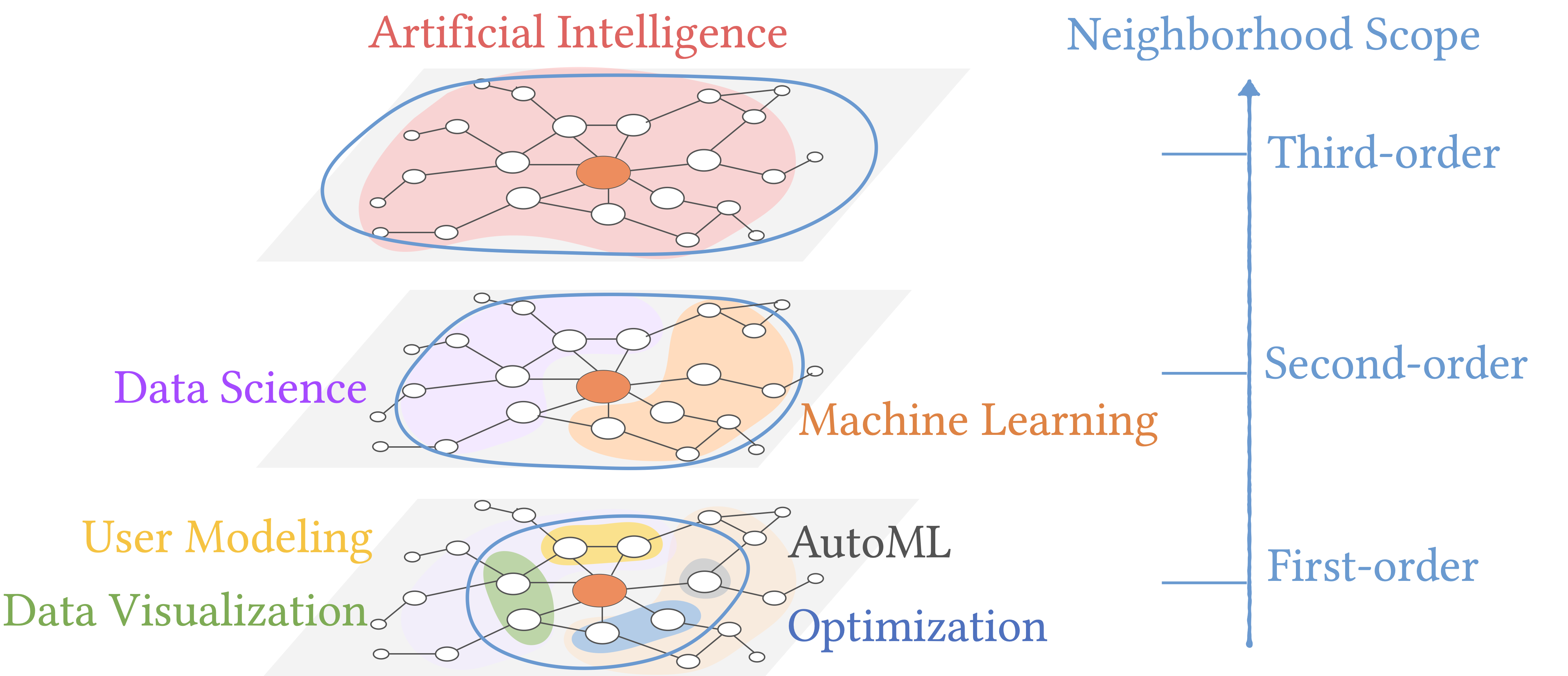}
\caption{An illustration of hierarchical grouping. The higher layer identifies more global features and contexts shared by larger groups of nodes.}
\label{fig:example} 
\vspace{-3mm}
\end{figure}

Graphs are pervasive due to their expressiveness of complex relations between entities in a variety of fields, ranging from social science \cite{rozemberczki2019multiscale, sen2008collective}, e-commerce \cite{yang2015defining, gong2020jnet} to biology \cite{li2017oncoppi} and many others. 
\textit{Graph embedding} is a technique to learn low-dimensional vector representations of nodes or subgraphs that preserve information about the original graph, such as topological structure and node properties. This technique has been effectively applied to a wide range of tasks, such as node classification \cite{liu2019single}, link prediction \cite{epasto2019single, lin2020graph}, and community detection \cite{cavallari2017learning}. 
Recent graph embedding methods achieved phenomenal success by exploiting the \textit{neighborhood structure} of each node.
Specifically, each node is embedded by aggregating  \textit{local features} from its neighboring nodes \cite{kipf2016semi}; then the node embedding is decoded to recover the node's \textit{structural context}, which is composed of neighbors that can be related to this node (e.g., reachable via random walks in the original graph  \cite{perozzi2014deepwalk}).

To effectively encode the local features and preserve the surrounding context, it is worth noting that \textit{latent hierarchical grouping of nodes} commonly exists in a graph's neighborhood structure.
Nodes tend to form groups by sharing common features and contexts; and the groups are organized in a hierarchical manner \cite{long2019hierarchical, clauset2008hierarchical, clauset2006structural}:
at the lower level, groups formed among immediate neighboring nodes identify more fine-grained features and contexts shared by these nodes; as we broaden the scope of neighborhood to reach higher-order neighbors, small groups at lower-levels are merged with more globally shared patterns and contexts extracted.

Figure \ref{fig:example} shows a motivating example on a citation graph to illustrate our intuition. The nodes denoting papers are connected by citation links.
The two layers in the figure concern different \textit{neighborhood scopes} indicated by the blue circle, which is defined by the order of neighborhood reachable from the red target node. 
From the bottom up, as we extend the neighborhood scope to include higher-order neighbors, small groups with more fine-grained meanings are merged to larger groups. 
For instance, if we inspect the first-order neighbors of the target node, we  may distinguish them relating to ``Optimization'' or ``AutoML'', because more subtle distinctions between them can be identified from their node features and direct citation links. But when we consider the higher-order neighborhood, we may view nodes previously from different fine-grained groups to be members of the same but more coarse-grained groups, such as ``Machine Learning'' vs., ``Data Science''.

The same insight has been investigated in social psychology, which suggests that related individuals work as \textit{superorganisms}  \cite{kesebir2012superorganism,johnson2010deconstructing} where each individual has a capacity to partly and flexibly manifest their properties in different groups, and new properties emerge when small groups are merged into larger ones. 
This inspires us to view graphs as living organisms, where a hierarchy of groups is essential to discover emergent relation among nodes under different contexts, and also to understand the influence of neighborhood both locally and globally.

Although the hierarchical grouping of graph nodes is ubiquitous and informative, little attention has been devoted to it. 
Previous works showed that modeling a node's membership to multiple groups is effective in refining the relatedness among nodes \cite{lin2020graph, long2020graph, duong2019unsupervised, park2020unsupervised}.
For example, the connection among nodes was profiled as a distribution of memberships, which is inferred from textual content \cite{lin2020graph} or extracted from random walks \cite{long2020graph} based on each node's neighborhood \cite{duong2019unsupervised, park2020unsupervised}.
However, these works simply modeled the grouping of nodes with a global mixture model, ignoring the hierarchy among groups. 
Other attempts trying to preserve hierarchical community structure were purely based on graph structure \cite{du2018galaxy, long2019hierarchical}, and ignored its inter-dependency with node features in the neighborhood structure.

Modeling nodes' membership to hierarchical groups establishes a new principle for graph embedding: 
\textit{nodes that are members of the same group should be embedded closely, and the learned node embeddings should reflect the nodes' membership to groups at different layers in the hierarchy.}
In light of this principle, we define two important properties of group hierarchy in a graph: 
1) the layer of the hierarchy controls the resolution of node groups, such that the higher-level layers capture global patterns shared by a broader neighborhood scope; 
2) an inclusive relation exists across layers, such that lower layer groups are merged into higher layer groups carrying more shared patterns from the bottom up in the hierarchy.


To realize such a group hierarchy when exploiting neighborhood structure, we propose a novel Graph embedding model with Hierarchical Attentive Memberships, abbreviated as \ourModel{}. 
In this model, we embed nodes and latent node groups to the same latent space, such that the affiliation of nodes toward groups can be inferred based on its context.
The critical design component is an aggregation function that attends neighboring nodes guided by both node and group embeddings jointly, such that both local and global context within the neighborhood scope are captured.
The node states generated for each aggregation layer are trained to recover context nodes at a certain granularity for the neighborhood scope of this layer.
We further incorporate a set of structural constraints on the inferred group memberships across layers of the hierarchy, such that the inclusive relation across layers is explicitly impose for a well-defined group hierarchy.

We conducted extensive experiments on a variety of graphs from citation networks to social networks, and demonstrated the effectiveness of the proposed model on node classification and link prediction tasks, against a list of state-of-the-art solutions.
We further visualized the learned node embeddings along with inferred memberships in each layer, and clear membership hierarchies were identified in real graphs, which demonstrated the potential of our solution in providing explainable graph embeddings.

%% file: related.tex
\section{Related Work}
\noindent\textbf{Hierarchical graph embedding.}
Recent years have witnessed numerous advances in deep architectures for learning graph embeddings,
among which Graph Convolutional Networks (GCNs) received the most attention  \cite{bruna2013spectral, defferrard2016convolutional, kipf2016semi, hamilton2017inductive}.
While most GCN models consider nodes as homogeneous, there are some efforts exploring the hierarchical grouping property of nodes in graphs. 
Hierarchical structure could be \textit{observed} in heterogeneous multi-dimensional graphs where nodes are grouped by different types of relations. In such cases, embeddings are learned to preserve both node-specific and group-shared information \cite{ma2018multi}. 
However, in most cases the hierarchical groups are \textit{latent}, and there are mainly two lines of studies addressing this challenge: 1) graph coarsening methods, and 2) spherical projection methods. 

\textit{Graph coarsening} strategies are proposed to obtain a series of successively simplified graphs capturing global patterns under different granularity \cite{chen2018harp, ying2018hierarchical, hu2019hierarchical, bianchi2020spectral, li2020structural, zhang2020enhanced}. Graphs are coarsened by a node and edge collapsing heuristic in \cite{chen2018harp}; however, it is performed as a pre-processing step thus is isolated from later model training. 
Graph pooling operation is proposed to learn an intermediate weight matrix as a soft group assignment, such that nodes are merged as hyper-nodes to coarsen graphs \cite{ying2018hierarchical, hu2019hierarchical}. Graph cut based clustering regularization is later introduced for graph pooling \cite{bianchi2020spectral}. However, such graph pooling methods lack necessary control on the weight matrix to maintain reasonable group hierarchy, such as the inclusive relation.
\textit{Spherical projection} methods embed low-level groups onto a spherical surface, the center of which represents the higher-level groups \cite{du2018galaxy, long2019hierarchical}. But they only consider node co-occurrences when constructing the hierarchy.
Hierarchical structure is also studied in the hyperbolic space \cite{nickel2017poincare}; but the learned embeddings in such space are difficult to be converted into the Euclidean vectors, which limits their applications.

\noindent\textbf{Multi-membership based graph embedding}.
Nodes are considered to have distinct memberships to different groups depending on its local context \cite{lin2020graph, park2020unsupervised, long2020graph, peixoto2014hierarchical, lei2020consistency}.
\citet{long2020graph} model node memberships to topic groups on random walks in the graph. The solution heavily relies on the random walks generated prior to the actual embedding, which results in fixed memberships that are independent from the training of embedding models. 
\citet{lin2020graph} associate each edge with multiple group channels, and propose a channel-aware attention mechanism to aggregate neighbor features based on their memberships. But external supervision from textual content in a graph is needed to infer the memberships.
There are some recent studies that dynamically assign  membership in each node based on a single context node \cite{duong2019unsupervised, park2020unsupervised}. The selection module is trained with node embedding via Gumbel-Softmax \cite{park2020unsupervised} or a graph cut loss \cite{duong2019unsupervised}, but group hierarchy is still ignored.

Though the effectiveness of multi-membership modeling has been verified in the aforementioned studies, the memberships are based on a flat set of groups with the same granularity \cite{duong2019unsupervised, park2020unsupervised, long2020graph}.
Our work proposes a more general model to capture nodes' membership for a hierarchy of groups under different granularity. The node groups at each layer of the hierarchy correspond to a certain neighborhood scope. It enables structured and automatic membership discovery for each node, which supports more comprehensive and accurate node embeddings.

%% file: method.tex
\begin{figure*}[tb] 
\centering
\includegraphics[width=7.0in]{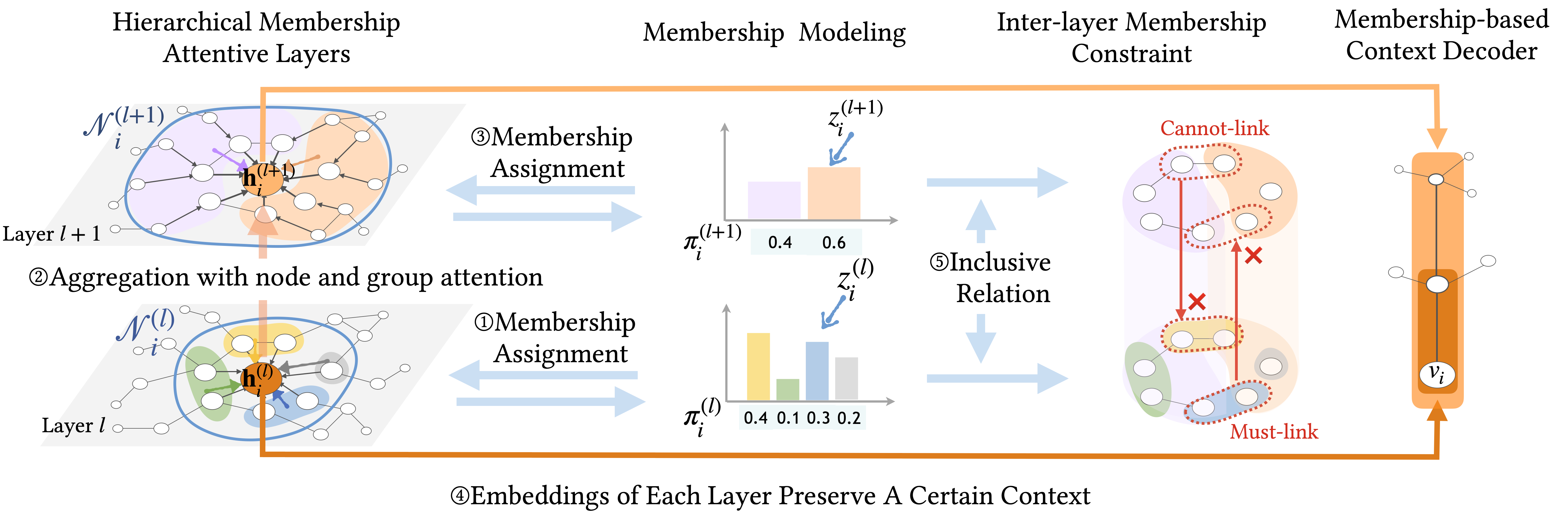}
\caption{Overview of \ourModel. In each layer of the aggregation operation, a group membership is firstly sampled for each node. Then the information from neighbors is attended by the inferred membership to generate node states for the next layer. The node states for each layer are learned by recovering the context within a certain neighborhood scope. Meanwhile, inter-layer constraints are introduced to inject the inclusive relation between membership assignments across layers.}
\label{fig:frame} 
\vspace{-2mm}
\end{figure*}

\section{Model: \ourModel}
In this section, we explain the detailed design of our model \ourModel\ as illustrated in Figure \ref{fig:frame}. Each node is modeled as a mixture of groups at each layer (Section \ref{sec:membership}). The hierarchical membership attentive layer aggregates information from neighboring nodes with both node-level and group-level attentions, such that nodes sharing local and global context are embedded closely (Section \ref{sec:layer}). Finally, the node embedding vectors on each layer are decoded by preserving the structural contexts within a certain neighborhood scope (Section \ref{sec:decoder}). We further impose structural constraints in the form of must-links and cannot-links to regularize the inferred memberships across layers (Section \ref{sec:regularization}).

\subsection{Membership Modeling}
\label{sec:membership}
Both the node groups and nodes' memberships are latent. We propose to infer nodes' affiliation to each group based on their embeddings. Then each time a group assignment is sampled for a node from its membership when it interacts with another node, adopting the intuition that the group membership of each node is \textit{context dependent} and its particular assignment is only manifested under a specific context.

Given graph $G=(V, E)$ with $V$ denoting a set of nodes and $E$ representing the edges, we assume that each node $v_i$ is associated with an embedding vector $\mathbf{h}_i\in\mathbb{R}^{d}$ that depicts the states of this node. 
As illustrated in Figure \ref{fig:example}, groups can be formed by nodes that share similar states, when we view node proximity in a certain neighborhood scope. 
We assume that each node is a mixture of $K$ groups with a corresponding group-membership distribution denoted by $\pi_i\in\Delta^{K}$, where $\Delta^{K}$ is a probability simplex over $K$ dimensions, i.e., $\forall k, \pi^k_i\ge0$ and $\sum^K_{k=1} \pi^{k}_{i}=1$.
To infer the latent membership of each node, we embed each group in the same space as node state denoted by $\phi_k\in\mathbb{R}^{d}$, and use $\Phi\in\mathbb{R}^{K\times d}$ to represent the matrix of embeddings of all $K$ groups.
Therefore, given a node state $\mathbf{h}_i$, we measure the node's affiliation to each group via  $\Phi\cdot\mathbf{h}_i\in\mathbb{R}^{K\times 1}$.
We then sample a group assignment for each node denoted as $z_i$, following the commonly adopted assumption that each node only manifests a single membership depending on the specific context \cite{airoldi2008mixed, park2020unsupervised}.
We summarize the procedure in the following steps.

\begin{itemize}
    \item For each node $v_i$:
    \begin{itemize}
        \item Draw its membership vector $\pi_i \sim \text{Dir}(\text{softmax}(\Phi\cdot\mathbf{h}_i))$
        \item Draw its group assignment $z_i\sim \text{Multi}(\pi_i)$
    \end{itemize}
\end{itemize}
where $\text{Dir}(\cdot)$ denotes the Dirichlet distribution and $\text{Multi}(\cdot)$ is the Multinomial distribution. 
Note that the hard assignment based on a categorical distribution is non-differentiable, which blocks the flow of gradients in later end-to-end optimization. 
We adopt the Gumbel-Softmax trick \cite{li2018deeper} to reparameterize the multinomial distribution and draw a one-hot assignment $z_i$ as follows:
\begin{equation*}
    z_i=\text{one-hot}(\text{argmax}_{k}[\pi_{i,k} + g]), g\sim\text{Gumbel}(0,1)
\end{equation*}
where $g$ is sampled from the standard Gumbel distribution.
The non-differentiable $\text{argmax}(\cdot)$ operation is further replaced with softmax to render the whole process differentiable:
\begin{align}
\label{eq:z}
    z_{i} \propto \text{exp}\big((\pi_{i,k}+g)/\tau\big) 
\end{align}
where $\tau$ is the temperature parameter to control the extent to which the output approximates the $\text{argmax}(\cdot)$  operation: As $\tau\rightarrow0$, samples from the Gumbel-Softmax distribution become one-hot.

Multiple sets of groups can be modeled when considering nodes with different scopes of neighborhood.
As shown in Figure \ref{fig:frame}, for layer $l+1$ concerning neighborhood scope $\mathcal{N}^{(l+1)}$, the target node marked by red is associated with a group-membership distribution $\pi_i^{(l+1)}$ obtained by group embeddings $\Phi^{(l+1)}\in\mathbb{R}^{K^{(l+1)}\times d^{(l+1)}}$ and node states $\mathbf{h}_{i}^{(l+1)}\in\mathbb{R}^{d^{(l+1)}}$. It is then assigned a new group $z_i^{(l+1)}$. 
Since the higher-level layer captures more coarse-grained groups covering larger neighborhood scope, the number of groups should become smaller, i.e., $K^{(l)} > K^{(l+1)}$.

\subsection{Hierarchical Membership Attentive Layers}
\label{sec:layer}
Our principle suggests that node proximity in the embedding space should reflect their closeness in the group hierarchy. 
Following this insight, we propose an aggregation function that attends neighboring nodes by both node-level and group-level relatedness. 

Formally, the aggregation operation of graph convolutional layer $l+1$ takes two inputs: 1) \textbf{node states} from the previous layer $l$, $\{\mathbf{h}_{1}^{(l)},\dots,\mathbf{h}^{(l)}_{|V|}\}$, where $\mathbf{h}_{i}^{(l)}\in\mathbb{R}^{d^{(l)}}$; 2) \textbf{group states} within the neighborhood scope $\mathcal{N}^{(l)}$, which is profiled by the matrix of group embeddings $\Phi^{(l)}$.
The layer $l+1$ aggregates information from neighborhood scope $\mathcal{N}^{(l)}$ and generates a new set of node states $\{\mathbf{h}^{(l+1)}_{1}, \dots,\mathbf{h}^{(l+1)}_{|V|}\}, \mathbf{h}'_{i}\in\mathbb{R}^{d^{(l+1)}}$. We stack multiple layers to capture the information from neighborhoods of different scopes, where the node states output by a lower layer are used as input to the layer above it. We denote the raw input node features as $\mathbf{h}^{(1)}$, the first-order neighbor scope as $\mathcal{N}^{(1)}$, and stack $L$ layers.

When encoding the target node $v_i$, we propose the following aggregation function to emphasize neighboring nodes that have similar states and belong to related groups:
\begin{equation}
\label{eq:attention}
    \mathbf{h}_{i}^{(l+1)}=\sigma\left(\frac{1}{M}\sum\nolimits^{M}_{m=1}\sum\nolimits_{j\in\mathcal{N}_{i}}\lambda^m_{ij}\alpha^m_{ij}\mathbf{W}^{(l+1),m}\mathbf{h}_{j}^{(l)}\right)
\end{equation}
where $\mathcal{N}_{i}$ is the immediate neighborhood of node $i$, $\{\textbf{W}^{(l+1),m}\in\mathbb{R}^{d^{(l+1)}\times d^{(l)}}\}^{M}_{m=1}$ is a set of state transformation matrices and $\sigma$ is a non-linear function. 
The states of neighboring nodes are re-weighed by both a group-level coefficient $\lambda_{ij}^{m}$ and a node-level coefficient $\alpha_{ij}^{m}$, which are calculated by multi-head attention \cite{velivckovic2017graph}. 
Specifically, we calculate the following attention weights with a shared transformation parameterized by $\mathbf{W}^{(l+1),m}$ for each head $m$:
\begin{align*}
    \lambda_{ij}^m &=\text{att}\left(\mathbf{W}^{(l+1),m}\Phi^{(l)}_{z^{(l)}_{i}}, \mathbf{W}^{(l+1),m}\Phi^{(l)}_{z^{(l)}_{j}}\right),\\ \alpha_{ij}^m &=\text{att}\left(\mathbf{W}^{(l+1),m}\mathbf{h}^{(l)}_i, \mathbf{W}^{(l+1),m}\mathbf{h}^{(l)}_j\right).
\end{align*}
Thus $\lambda_{ij}^{m}$ captures the global relatedness between the two nodes' assigned groups, while $\alpha_{ij}^{m}$ measures the local relatedness in terms of node states. The attention function $\text{att}(\cdot, \cdot)$ \cite{bahdanau2014neural} can be expressed as follows by normalizing over all nodes within the neighborhood: 
\begin{equation*}
    \text{att}(\mathbf{p}_i, \mathbf{p}_j)=\frac{\exp{\big(\text{LeakyReLU}(\textbf{a}^{\top}[\mathbf{p}_i\|\mathbf{p}_j])\big)}}{\sum_{j'\in\mathcal{N}_{i}}\exp{(\text{LeakyReLU}\big(\mathbf{a}^{\top}[\mathbf{p}_i\|\mathbf{p}_j'])\big)}}
\end{equation*}
where $\|$ is the concatenation operation over two vectors, and $\mathbf{a}\in\mathbb{R}^{2d^{(l)}}$ is the weight vector of a linear transformation.

The combination of group-level and node-level attention is the key to realizing our principle: if two nodes have similar node states measured by $\alpha$, and belong to related groups indicated by $\lambda$, more information should be passed between them; and thus they are encoded closer in the embedding space.
As a result, the relatedness of nodes with respect to the hierarchy of groups is preserved in the learnt embedding space. 

\subsection{Membership-based Context Decoder}
\label{sec:decoder}
Our proposed principle argues that node embeddings should preserve the group hierarchy, which requires us to capture node proximity at each layer of the hierarchy.
This can be achieved by aligning the neighborhood scope when aggregating information and decoding the context. The former is achieved by our membership-aware attentive layer introduced above, and now 
we introduce our membership-based context decoder to preserving graph hierarchy.

Given a target node $v_i$, the context summarizes its surrounding nodes when it manifests a certain membership.
We introduce a trainable membership-based context vector $\mathbf{Q}^{(l)}_{j,z_{i}^{(l)}}\in\mathbb{R}^{d^{(l)}}$ to encode each node $v_j$ that constitutes the context of the target node $v_i$ within the neighborhood scope $\mathcal{N}_i^{(l)}$.
This context vector can be decoded by maximizing the likelihood of observing the context nodes given a target node, defined by the following skip-gram based objective \cite{park2020unsupervised}:
\begin{align}
\label{eq:loss-context}
     \mathcal{L}^{(l)}_{context}
    = \sum_{v_i\in V}\sum_{j\in\mathcal{N}_i^{(l)}}&-\log p\big(v_j|v_i, z_i^{(l)}\big)\\
    = \sum_{v_i\in V}\sum_{j\in\mathcal{N}_i^{(l)}}&-\log\Big(\sigma(\mathbf{h}^{(l)\top}_{i}\cdot\mathbf{Q}^{(l)}_{j,z_i^{(l)}})\Big)\nonumber\\ 
    &- \mathbb{E}_{j_n\sim{\mathcal{\bar{N}}^{(l)}_i}}\log\Big(\sigma(-\mathbf{h}^{(l)\top}_{i}\cdot\mathbf{Q}^{(l)}_{j_n,z_i^{(l)}})\Big).
\nonumber
\end{align}
The context vector $\mathbf{Q}^{(l)}_{j,z_{i}^{(l)}}$ depends on the group assignment $z_i^{(l)}$ of the target node $v_i$, which aligns with our intuition that a single membership is manifested in a given context. 
$\mathcal{\bar{N}}^{(l)}_{i}$ denotes the set of node outside the immediate neighborhood of $v_i$. 
This objective represents the context reconstruction error, such that nodes that frequently co-occur within scope $\mathcal{N}^{(l)}$ should be pushed closer in the embedding space.
We use negative sampling to construct $\mathcal{\bar{N}}^{(l)}_{i}$ for efficient calculation. 

As illustrated in Figure \ref{fig:frame}, node state generated by each layer is decoded to recover the context composed of the neighborhood with corresponding scope (indicated by orange arrows). By aligning the neighborhood scope in the aggregation layer and the decoded context, we capture the anticipated property of graph hierarchy that each layer controls the resolution of groups.

\subsection{Inter-layer Membership Constraints}
\label{sec:regularization}
An inclusive relation exists in the hierarchy of groups to depict the emergence of groups bottom up. We leverage this relation to explicitly introduce inter-layer constraints to regularize the modeling of latent group memberships of nodes.

We define two sets of constraints: 
1) \textbf{must-link} denoted by $\mathcal{M}^{(l+1)}=\{(v_{i}, v_{j}): z_i^{(l)}=z_j^{(l)}\}$, which implies that nodes $v_i$ and $v_j$ should be members of the the same group in the higher layer $l+1$, as they already belong to the same group in the lower layer;
2) \textbf{cannot-link} denoted by $\mathcal{C}^{(l)}=\{(v_{i}, v_{j}): z_i^{(l+1)}\neq z_j^{(l+1)}\}$, which suggests that these two nodes should belong to different groups in the lower layer $l$, since they belong to different groups in the higher layer.
We then introduce the following regularization term to penalize the violation of must-link and cannot-link constraints:
\begin{equation}
    \mathcal{L}_{reg}^{(l)}=\sum_{(v_i,v_j)\in\mathcal{M}^{(l+1)}}\gamma\cdot\mathbbm{1}\Big(z_i^{(l+1)}\neq z_j^{(l+1)}\Big)+\sum_{(v_i,v_j)\in\mathcal{C}^{(l)}}\beta\cdot\mathbbm{1}\Big(z^{(l)}_i=z_j^{(l)}\Big)
\label{eq:loss-reg}
\end{equation}
where $\mathbbm{1}(\cdot)$ is the indicator function. The strength of penalty is controlled by $\gamma$ and $\beta$ respectively.

The node pairs marked by red circles in Figure \ref{fig:frame} illustrate the purpose of the inter-layer membership constraints, where two pairs of nodes are penalized for violating the must-link and cannot-link requirements on membership assignments.
This regularization ensures the dependency between memberships across layers, such that node proximity maintains the consistency in the hierarchy.

Applying the inter-layer constraints to the loss of recovering membership-based contexts with different neighborhood scopes, we obtain the final optimization objective as follows to learn the node states generated on each aggregation layer in \ourModel{}:
\begin{equation}
\label{eq:final-loss}
    \mathcal{L}=\sum\nolimits_{l=1}^{L}\mathcal{L}_{context}^{(l)}+\sum\nolimits_{l=1}^{L-1}\mathcal{L}_{reg}^{(l)}
\end{equation}
To prepare the embeddings for use on downstream tasks, such as node classification, we concatenate the layer-wise node states to obtain a final embedding of each node, which can be further fine-tuned by introducing a task-specific loss to Eq \eqref{eq:final-loss}. 

%% file: experiment.tex
\section{Evaluation}

\begin{table}
  \caption{Statistics of evaluation graph benchmark datasets.}
 \vspace{-2mm}
  \label{tab:data}
  \begin{tabular}{crrcc}
    \toprule
    Dataset & \#Node & \#Edge & \#Class & Clustering Coef. \\
    \midrule
    Cora & $2,708$ & $5,429$ & $7$ & $0.241$ \\
    Citeseer & $3,327$ & $4,732$ & $6$ & $0.141$ \\
    Pubmed & $19,717$ & $44,338$ & $3$ & $0.060$ \\
    Facebook & $22,470$ & $171,002$ & $4$ & $0.360$ \\
    Youtube & $1,138,499$ & $2,990,443$ & $47$ & $0.080$ \\
    Amazon & $2,449,029$ & $123, 718, 280$ & $47$ & $0.419$ \\ 
    \bottomrule
  \end{tabular}
  \vspace{-2mm}
\end{table}

We evaluated \ourModel\ on two popular tasks, node classification and link prediction, to verify its effectiveness in preserving node property and graph structure (Section \ref{sec:quantitative}). In the qualitative analysis, we mapped the learned node embeddings to a 2-D space to demonstrate the membership inferred by \ourModel{}, which verified the effectiveness of the learned group-membership distributions in discovering nodes at the boundary of groups (Section \ref{sec:qualitative}).
Finally, we analyzed \ourModel{} via a comprehensive ablation study which verified the effectiveness of membership attentive layers and inter-layer membership regularization (Section \ref{sec:model-analysis}).

\subsection{Experiment Setup}
\noindent\textbf{$\bullet$ Datasets.}
We included six public datasets for our evaluation, ranging from academic citation networks to large-scale social networks. The citation network datasets, Cora, Citeseer and Pubmed \cite{sen2008collective}, contain research papers as nodes and citation links as edges. The Facebook dataset \cite{rozemberczki2019multiscale} represents official Facebook homepages as nodes and mutual likes between them as edges. The Youtube dataset \cite{tang2009scalable} includes users as nodes and co-subscription relations as edges. The Amazon graph \cite{chiang2019cluster} denotes products as nodes which are connected by edges if purchased together. Table \ref{tab:data} shows detailed statistics of the datasets, and the clustering coefficient measures the degree to which nodes tend to be clustered together.

\noindent\textbf{$\bullet$ Baselines.} 
The proposed \ourModel\ is compared against a rich collection of graph embedding models: 
1) \textbf{GraphSage} \cite{hamilton2017inductive} uniformly passes information through edges without  neighborhood-based attention.
2) \textbf{GAT} \cite{velivckovic2017graph} incorporates attention mechanism on node states to reweigh neighboring nodes when aggregating their information.
3) \textbf{GraphSTONE} \cite{long2020graph} is a multi-membership baseline, but ignores the hierarchy of groups. It extracts structural patterns as groups from random walks, and uses a topic model to infer group membership for reweighing neighbor nodes.
4) \textbf{DeepMinCut} \cite{duong2019unsupervised} is a multi-membership  baseline which derives nodes' membership assignments by minimizing a graph cut loss.
5) \textbf{asp2vec} \cite{park2020unsupervised} is a multi-membership baseline which dynamically assigns each node a membership based on its context in random walk.
6) \textbf{H-GCN} \cite{hu2019hierarchical} is a hierarchical embedding baseline, which coarsens graphs by learning a soft membership assignment matrix for each aggregation layer, and then produces node-level embeddings by refining the layers under node classification loss. 
7) \textbf{GNE \cite{du2018galaxy}} is a hierarchical embedding baseline based on spherical projection, which projects lower-level groups to a sphere with the center representing the merged higher-level groups.
8) \textbf{SpaceNE \cite{long2019hierarchical}} projects groups into different subspaces whose dimensionalities reflect the hierarchy, such that groups in lower-dimension subspace can be merged in higher-dimension subspace. 

\noindent\textbf{$\bullet$ Experiment settings.}
We set the node embedding size $d=128$, and use $L=2$ aggregation layers for all GCN-based methods. Each layer in \ourModel{} has its dimension set to $d_1=d_2=d/L=64$.
The models are trained in a mini-batch manner following \cite{hamilton2017inductive}.
For each node in a batch, we sample $S_1=S_2=25$ neighbors for each layer. 
To sample node pairs for skip-gram optimization, we conduct random walks from each node 50 times with a window size set to 2. We set the number of negative samples equal to the number of positive examples.
The other parameters of baselines are set to their optimal values as suggested in their original papers.
All the results are reported based on five-fold cross-validation.

\noindent\textbf{$\bullet$ Model complexity.}
Compared with commonly used GCNs, we only introduced two sets of additional parameters to model group membership: the membership embeddings $\Phi^{(l)}\in\mathbb{R}^{K^{(l)}\times d^{(l-1)}}$ for each aggregation layer, and the context node embeddings $\mathbf{Q}^{(l)}\in\mathbb{R}^{K^{(l)}\times d^{(l)}}$ for the membership-based context decoder. In general, GCNs are equipped with a weight matrix of state transformation $\mathbf{W}^{(l)}\in\mathbb{R}^{d^{(l)}\times d^{(l-1)}}$ for each layer. Since the number of groups $K$ is usually set smaller than the embedding dimension $d$, we did not significantly increase model complexity in \ourModel. 

\begin{table*}[t]
    \setlength{\tabcolsep}{0.5em}
	\centering
	\caption{Performance comparisons of node classification task under different metrics.}
	\label{tab:node}
	\renewcommand{\arraystretch}{0.8} 
    \linespread{0.8}
	\begin{tabular}{>{\centering}lccccccccc}
	\toprule
	Dataset & Metric & GraphSage & GAT & H-GCN & GraphSTONE &
	GNE & SpaceNE & DeepMinCut & \textbf{\ourModel} \\
	\midrule
	\multirow{3}{*}{Cora} & 
	Accuracy & 
	$0.812{\scriptstyle \pm0.004}$ & 
	$0.829{\scriptstyle \pm0.003}$ & 
	$0.845{\scriptstyle \pm0.003}$ & 
	$0.823{\scriptstyle \pm0.006}$ & 
	$0.787{\scriptstyle \pm0.015}$ &
	$0.796{\scriptstyle \pm0.011}$ &
	$0.840{\scriptstyle \pm0.003}$ & 
	$\bm{0.853}{\scriptstyle \pm0.004}$ \\
	& Micro-F1 &
	$0.823{\scriptstyle \pm0.004}$ &
	$0.834{\scriptstyle \pm0.003}$ &
	$0.854{\scriptstyle \pm0.004}$ &
	$0.831{\scriptstyle \pm0.005}$ &
	$0.782{\scriptstyle \pm0.009}$ &
	$0.788{\scriptstyle \pm0.008}$ &
	$0.840{\scriptstyle \pm0.004}$ &
	$\bm{0.860}{\scriptstyle \pm0.003}$ \\
	& Macro-F1 &
	$0.785{\scriptstyle \pm0.003}$ &
	$0.801{\scriptstyle \pm0.003}$ &
	$0.818{\scriptstyle \pm0.003}$ &
	$0.798{\scriptstyle \pm0.005}$ &
	$0.755{\scriptstyle \pm0.0011}$ &
	$0.746{\scriptstyle \pm0.009}$ &
	$0.823{\scriptstyle \pm0.004}$ &  
	$\bm{0.824}{\scriptstyle \pm0.003}$ \\
	\midrule
	\multirow{3}{*}{Citeseer} &
	Accuracy &
	$0.703{\scriptstyle \pm0.005}$ &
	$0.723{\scriptstyle \pm0.004}$ &
	$0.724{\scriptstyle \pm0.003}$ &
	$0.720{\scriptstyle \pm0.005}$ &
	$0.676{\scriptstyle \pm0.0016}$ &
	$0.679{\scriptstyle \pm0.0013}$ &
	$0.723{\scriptstyle \pm0.004}$ &
	$\bm{0.727}{\scriptstyle \pm0.004}$ \\
	& Micro-F1 &
	$0.756{\scriptstyle \pm0.004}$ &
	$0.772{\scriptstyle \pm0.004}$ &  
	$0.781{\scriptstyle \pm0.003}$ &
	$0.768{\scriptstyle \pm0.006}$ &
	$0.722{\scriptstyle \pm0.0011}$ &
	$0.725{\scriptstyle \pm0.0010}$ &
	$0.723{\scriptstyle \pm0.005}$ &
	$\bm{0.783}{\scriptstyle \pm0.003}$ \\
	& Macro-F1 &
	$0.724{\scriptstyle \pm0.005}$ & 
	$0.739{\scriptstyle \pm0.004}$ & 
	$\bm{0.748}{\scriptstyle \pm0.003}$ &
	$0.739{\scriptstyle \pm0.005}$ &
	$0.706{\scriptstyle \pm0.012}$ &
	$0.710{\scriptstyle \pm0.010}$ &
	$0.682{\scriptstyle \pm0.006}$ &
	$0.746{\scriptstyle \pm0.004}$\\
	\midrule
	\multirow{3}{*}{Pubmed} &
	Accuracy &
	$0.788{\scriptstyle \pm0.005}$ & 
	$0.797{\scriptstyle \pm0.004}$ &
	$0.797{\scriptstyle \pm0.004}$ &  
	$0.794{\scriptstyle \pm0.006}$ &
	$0.749{\scriptstyle \pm0.011}$ &
	$0.756{\scriptstyle \pm0.011}$ &
	$0.723{\scriptstyle \pm0.005}$ &
	$\bm{0.812}{\scriptstyle \pm0.005}$ \\
	& Micro-F1 &
	$0.802{\scriptstyle \pm0.004}$ & 
	$0.813{\scriptstyle \pm0.004}$ &  
	$0.817{\scriptstyle \pm0.005}$ &
	$0.809{\scriptstyle \pm0.006}$ &
	$0.763{\scriptstyle \pm0.009}$ &
	$0.775{\scriptstyle \pm0.010}$ &
	$0.723{\scriptstyle \pm0.005}$ &
	$\bm{0.824}{\scriptstyle \pm0.005}$ \\
	& Macro-F1 &
	$0.794{\scriptstyle \pm0.004}$ &
	$0.801{\scriptstyle \pm0.005}$ & 
	$0.813{\scriptstyle \pm0.005}$ & 
	$0.804{\scriptstyle \pm0.007}$ &
	$0.755{\scriptstyle \pm0.003}$ &
	$0.762{\scriptstyle \pm0.011}$ &
	$0.712{\scriptstyle \pm0.011}$ &
	$\bm{0.819}{\scriptstyle \pm0.004}$ \\
	\midrule
	\multirow{3}{*}{Facebook} &
	Accuracy & 
	$0.875{\scriptstyle \pm0.006}$ &
	$0.894{\scriptstyle \pm0.007}$ & 
	$0.901{\scriptstyle \pm0.006}$ & 
	$0.905{\scriptstyle \pm0.008}$ &
	$0.864{\scriptstyle \pm0.017}$ &
	$0.848{\scriptstyle \pm0.018}$ &
	$0.876{\scriptstyle \pm0.007}$ &
	$\bm{0.918}{\scriptstyle \pm0.006}$ \\
	& Micro-F1 &
	$0.894{\scriptstyle \pm0.005}$ &
	$0.907{\scriptstyle \pm0.006}$ &  
	$0.915{\scriptstyle \pm0.006}$ & 
	$0.918{\scriptstyle \pm0.007}$ & 
	$0.867{\scriptstyle \pm0.014}$ &
	$0.852{\scriptstyle \pm0.012}$ &
	$0.879{\scriptstyle \pm0.006}$ &
	$\bm{0.930}{\scriptstyle \pm0.006}$ \\
	& Macro-F1 &
	$0.889{\scriptstyle \pm0.005}$ &
	$0.905{\scriptstyle \pm0.005}$ &
	$0.909{\scriptstyle \pm0.006}$ & 
	$0.912{\scriptstyle \pm0.007}$ & 
	$0.853{\scriptstyle \pm0.015}$ &
	$0.846{\scriptstyle \pm0.014}$ &
	$0.865{\scriptstyle \pm0.006}$ & 
	$\bm{0.924}{\scriptstyle \pm0.006}$ \\
	\midrule
	\multirow{3}{*}{Youtube} &
	Accuracy &
	$0.732{\scriptstyle \pm0.004}$ &
	$0.745{\scriptstyle \pm0.004}$ &
	$0.744{\scriptstyle \pm0.005}$ &
	$0.739{\scriptstyle \pm0.006}$ &
	$0.713{\scriptstyle \pm0.017}$ &
	$0.722{\scriptstyle \pm0.012}$ &
	$0.742{\scriptstyle \pm0.007}$ &
	$\bm{0.755}{\scriptstyle \pm0.005}$ \\
	& Micro-F1 &
	$0.775{\scriptstyle \pm0.004}$ &
	$0.787{\scriptstyle \pm0.005}$ &
	$0.782{\scriptstyle \pm0.005}$ &
	$0.784{\scriptstyle \pm0.006}$ &
	$0.753{\scriptstyle \pm0.016}$ &
	$0.759{\scriptstyle \pm0.014}$ &
	$0.767{\scriptstyle \pm0.007}$ &
	$\bm{0.795}{\scriptstyle \pm0.005}$ \\
	& Macro-F1 &
	$0.696{\scriptstyle \pm0.004}$ &
	$0.710{\scriptstyle \pm0.005}$ &
	$0.708{\scriptstyle \pm0.005}$ & 
	$0.711{\scriptstyle \pm0.005}$ & 
	$0.677{\scriptstyle \pm0.014}$ &
	$0.681{\scriptstyle \pm0.011}$ &
	$0.701{\scriptstyle \pm0.006}$ & 
	$\bm{0.721}{\scriptstyle \pm0.006}$ \\
	\midrule
	\multirow{3}{*}{Amazon} & 
	Accuracy & 
	$0.659{\scriptstyle \pm0.005}$ &
	$0.674{\scriptstyle \pm0.005}$ & 
	$0.652{\scriptstyle \pm0.006}$ & 
	$0.662{\scriptstyle \pm0.007}$ &
	$0.612{\scriptstyle \pm0.013}$ &
	$0.632{\scriptstyle \pm0.010}$ &
	$0.667{\scriptstyle \pm0.005}$ &
	$\bm{0.686}{\scriptstyle \pm0.006}$ \\
	& Micro-F1 &
	$0.727{\scriptstyle \pm0.005}$ &
	$0.743{\scriptstyle \pm0.005}$ &
	$0.731{\scriptstyle \pm0.006}$ &
	$0.733{\scriptstyle \pm0.007}$ &
	$0.687{\scriptstyle \pm0.009}$ &
	$0.703{\scriptstyle \pm0.010}$ &
	$0.705{\scriptstyle \pm0.005}$ &
	$\bm{0.742}{\scriptstyle \pm0.005}$ \\
	& Macro-F1 &
	$0.297{\scriptstyle \pm0.004}$ &
	$0.325{\scriptstyle \pm0.005}$ &
	$0.295{\scriptstyle \pm0.005}$ &
	$0.305{\scriptstyle \pm0.006}$ &
	$0.254{\scriptstyle \pm0.005}$ &
	$0.267{\scriptstyle \pm0.011}$ &
	$0.303{\scriptstyle \pm0.008}$ &
	$\bm{0.342}{\scriptstyle \pm0.005}$ \\
	\bottomrule
	\end{tabular}
\end{table*}

\subsection{Quantitative Analysis}
\label{sec:quantitative}
We evaluate \ourModel\ on node classification and link prediction tasks, and the improved performance on both tasks verifies the effectiveness of modeling the hierarchical group membership.

\noindent\textbf{$\bullet$ Node Classification.}
The node embeddings are fed as features to predict the node labels. Recall that each aggregation layer of \ourModel\ produces a vector for each node to encode neighbor information within a given scope. We concatenate these vectors from all layers to serve the node classification task. The composed embedding is trained in a multitask manner by joining the classification loss with Eq  \eqref{eq:final-loss}.

Table \ref{tab:node} summarizes the performance of \ourModel\ against baseline models with \textit{accuracy, micro- and macro- F1 score} metrics. 
We can clearly observe that \ourModel\ consistently outperformed baselines on these datasets. Compared with GraphSage and GAT which do not model group membership, \ourModel\ achieved a significant improvement. This demonstrates the importance of modeling the latent group structure in graphs.
GraphSTONE discovered global structures by summarizing random walk patterns; but \ourModel\ still outperformed it, which proves the effectiveness of inferring group membership in an end-to-end fashion.
DeepMinCut modeled global structures of the graph via graph cut, but it was less effective than our method because node features and the grouping hierarchy were ignored in DeepMinCut.
H-GCN proposed successive pooling operations which captured the grouping hierarchy, but lacked the control on the group assignments to calibrate the structure across layers.
GNE and SpaceNE recursively merged lower-level groups into higher-level groups in an unsupervised manner and ignored node features, thus gave worse performance. 

\noindent\textbf{$\bullet$ Link Prediction.} The task is to predict the linkage between two nodes via the similarity of their embeddings.
We formulate the task as a ranking problem to retrieve linked nodes from a candidate set with negative (irrelevant) nodes. 
We utilize \textit{Area under the ROC Curve (AUC)} and \textit{scaled mean reciprocal rank (MRR)} metrics to demonstrate how effectively the models can rank real neighbors at higher positions. 
Since the links between nodes correspond to the context defined by the first-order neighborhood, we use the embedding vectors generated by the first layer in \ourModel{} to make the prediction. 

\begin{figure*}[t]
\centering
\includegraphics[width=6.8in]{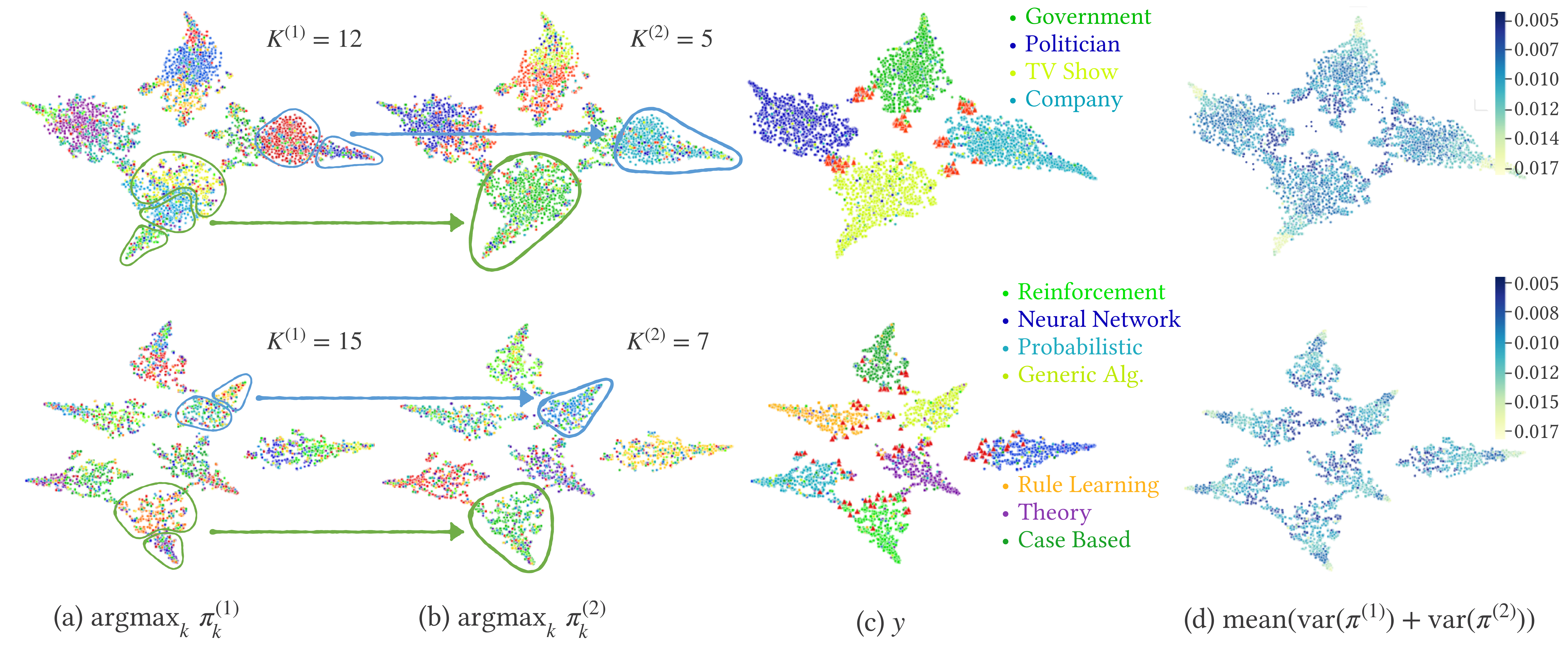}
\caption{Visualization of the learned embeddings on Facebook (upper) and Cora (lower) graph. Red triangles denote the boundary nodes with small variance of $\pi$. The rest colors denote the ground-truth classes ($Y$) or sampled group assignment across two layers ($z^{(1)}, z^{(2)}$) on each node.}
\label{fig:embed} 
\vspace{-2mm}
\end{figure*}

Table \ref{tab:link} summarizes the results of the link prediction task. 
GraphSAGE used two hidden layers to aggregate second-order neighborhood, which is empirically better than using one layer in it. In contrast, despite using only the output of a single hidden layer, \ourModel\ still outperformed GraphSAGE, which strongly suggests the effectiveness of our group membership modeling in capturing global patterns. GraphSTONE and asp2vec modeled multiple memberships, but ignored the inherent hierarchy concerning different granularity of contexts, and thus were still outperformed by \ourModel.

\begin{table}[t]
    \setlength{\tabcolsep}{0.3em}
	\centering
	\small
	\caption{Performance on link prediction task.}
	\label{tab:link}
	\renewcommand{\arraystretch}{0.8} 
    \linespread{0.8}
	\begin{tabular}{lccccc}
	\toprule
	Dataset & Metric & GraphSage & GraphSTONE & asp2vec & \textbf{\ourModel} \\
	\midrule
	\multirow{2}{*}{Cora} & 
	AUC &
	$0.849{\scriptstyle \pm0.018}$ &
	$0.858{\scriptstyle \pm0.015}$ &
	$0.865{\scriptstyle \pm0.021}$ &
	$\bm{0.869}{\scriptstyle \pm0.016}$ \\
	& MRR &
	$0.674{\scriptstyle \pm0.019}$ &
	$0.680{\scriptstyle \pm0.022}$ &
	$0.683{\scriptstyle \pm0.023}$ &
	$\bm{0.692}{\scriptstyle \pm0.018}$ \\
	\midrule
	\multirow{2}{*}{Citeseer} & 
	AUC &
	$0.935{\scriptstyle \pm0.015}$ &
	$0.944{\scriptstyle \pm0.017}$ &
	$0.944{\scriptstyle \pm0.015}$ &
	$\bm{0.957}{\scriptstyle \pm0.016}$ \\
	& MRR &
	$0.762{\scriptstyle \pm0.018}$ &
	$0.768{\scriptstyle \pm0.017}$ &
	$0.741{\scriptstyle \pm0.013}$ &
	$\bm{0.778}{\scriptstyle \pm0.015}$ \\
	\midrule
	\multirow{2}{*}{Pubmed} & 
	AUC &
	$0.953{\scriptstyle \pm0.017}$ &
	$\bm{0.962}{\scriptstyle \pm0.020}$ &
	$0.927{\scriptstyle \pm0.014}$ &
	$0.959{\scriptstyle \pm0.015}$ \\
	& MRR &
	$0.886{\scriptstyle \pm0.019}$ &
	$0.902{\scriptstyle \pm0.017}$ &
	$0.841{\scriptstyle \pm0.018}$ &
	$\bm{0.906}{\scriptstyle \pm0.016}$ \\
	\midrule
	\multirow{2}{*}{Facebook} & 
	AUC &
	$0.954{\scriptstyle \pm0.019}$ &
	$0.965{\scriptstyle \pm0.020}$ &
	$\bm{0.969}{\scriptstyle \pm0.016}$ &
	$0.966{\scriptstyle \pm0.018}$ \\
	& MRR &
	$0.832{\scriptstyle \pm0.016}$ &
	$0.844{\scriptstyle \pm0.022}$ &
	$0.842{\scriptstyle \pm0.024}$ &
	$\bm{0.855}{\scriptstyle \pm0.020}$ \\
	\midrule
	\multirow{2}{*}{Youtube} & 
	AUC &
	$0.757{\scriptstyle \pm0.014}$ &
	$0.763{\scriptstyle \pm0.016}$ &
	$0.750{\scriptstyle \pm0.013}$ &
	$\bm{0.768}{\scriptstyle \pm0.014}$ \\
	& MRR &
	$0.594{\scriptstyle \pm0.022}$ &
	$0.601{\scriptstyle \pm0.18}$ &
	$0.587{\scriptstyle \pm0.020}$ &
	$\bm{0.608}{\scriptstyle \pm0.019}$ \\
	\midrule
	\multirow{2}{*}{Amazon} & 
	AUC &
	$0.792{\scriptstyle \pm0.015}$ &
	$0.808{\scriptstyle \pm0.013}$ &
	$0.818{\scriptstyle \pm0.016}$ &
	$\bm{0.824}{\scriptstyle \pm0.014}$ \\
	& MRR &
	$0.579{\scriptstyle \pm0.017}$ &
	$0.584{\scriptstyle \pm0.015}$ &
	$0.589{\scriptstyle \pm0.022}$ &
	$\bm{0.593}{\scriptstyle \pm0.015}$ \\
	\bottomrule
	\end{tabular}
\end{table}

\subsection{Proof-of-Concept Visualization}
\label{sec:qualitative}
To analyze the quality of the jointly learned node embeddings and groups from \ourModel, we use the t-SNE algorithm to project the composed node embeddings to a 2-D space, and visualize Facebook graph in the first row and Cora graph in the second row in Figure \ref{fig:embed}. 
The node color has different meanings across the columns. 
In column (a) and (b), the color denotes the group assignment of each node $v_i$ with the largest affiliation strength in the first and second layer, i.e., $\text{argmax}_{k}\pi^{(1)}_{i,k}$ and $\text{argmax}_k\pi^{(2)}_{i,k}$, respectively. 
In column (c), the color shows the ground-truth label with its definition listed aside. 
In column (d), we calculate the average variance on $\pi^{(1)}$ and $\pi^{(2)}$ for each node to measure its degree of concentration over groups, and the color reflects the conentration: a darker color means a lower degree of concentration, which suggests that the node's affiliation to different groups is evenly distributed. 
Meanwhile, the nodes with the lowest concentration degree are highlighted by red triangles in column (c) for the purpose of illustration.

\begin{figure} 
\centering
\includegraphics[width=3.6in]{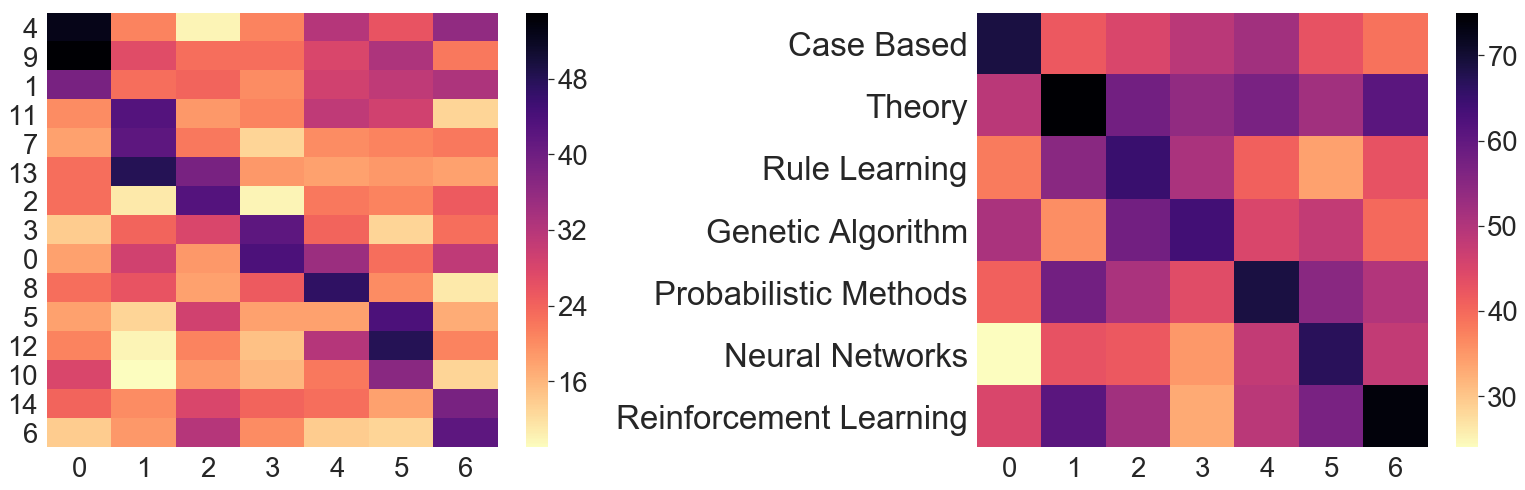}
\caption{Heatmaps about the degree of correlation between group assignments $z^{(1)}, z^{(2)}$ (left) and between $z^{(2)}, y$ (right) on Cora, measured by their concurrency on nodes.}
\label{fig:z} 
\vspace{-2mm}
\end{figure}

To better demonstrate the correlation between the learned groups and inferred memberships across layers and the ground-truth labels, we also visualize their co-occurrence matrices on Cora dataset, shown in Figure \ref{fig:z}. 
In the left heatmap, for each entry indexed by row $i$ and column $j$, we calculate the number of nodes that are concurrently assigned with $z^{(1)}=i$ in the first layer and $z^{(2)}=j$ in the second layer. In the right heatmap, we count the number of nodes that have ground-truth label $y=i$ and are assigned to group $z^{(2)}=j$ in the second layer. The color denotes the number of nodes satisfying those respective assignments, thereby reflecting the degree of correlation between $z^{(1)}, z^{(2)}$ and $y$.

The visualizations in Figure \ref{fig:embed} and Figure \ref{fig:z} together demonstrate two intriguing properties of \ourModel\ as discussed below.

\noindent\textbf{$\bullet$ A hierarchy of groups is captured.} Comparing column (a) and (b) of Figure \ref{fig:embed}, we can clearly observe different node groups are captured. And more interestingly, a hierarchy of groups is automatically discovered: in the highlighted circles, different groups of nodes shown in column (a) are merged to form larger groups shown in column (b). The trend of merging groups is a clear manifestation of the desired group hierarchy, where coarse-grained properties shared by a larger group of nodes are captured when we aggregate information from lower-level groups with fine-grained properties. 
Comparing column (b) and (c), we show that the learned node membership is also well aligned with node labels, which suggests that node label as a comprehensive signal to distinguish nodes is captured by the hierarchical group modeling. 
The correlation among $z^{(1)}, z^{(2)}$ and $y$ reported in Figure \ref{fig:z} also supports our argument. From the diagonals with large number of nodes, we observe a clear merging structure, where multiple entries of $z^{(1)}$ frequently co-occur with one specific entry in $z^{(2)}$. This suggests a trend of group merging from lower to higher layers. 

\noindent\textbf{$\bullet$ Our membership modeling discovers nodes at the boundary of groups.}
Recall that we calculate the variance of $\pi$ to measure the concentration of nodes' group affiliation. 
A low variance means that $\pi$ is flat such that the node's affiliation to different groups is evenly distributed. In other words, such nodes have no strong ties to any group and therefore reside at the boundary of circles \cite{abu2019mixhop}.  
In column (c) of Figure \ref{fig:embed}, we observe that the nodes marked by red triangles with the lowest variance are indeed located at the edge of different groups. 
The color gradient in Column (d) clearly demonstrates the coherence between the concentration degree over $\pi$ and the position of nodes in groups: the concentration decreases as we view from the center of a group to its boundary. 
This shows that modeling group membership encodes global structure, thereby endowing the learned embeddings with collective patterns in addition to the local pairwise proximity between nodes. 

\begin{table}[t]
    \setlength{\tabcolsep}{0.3em}
	\centering
	\small
	\caption{Ablation study of three model variants on node classification task. The results are performance gap between each variant and the complete \ourModel.
	}
	\label{tab:ablation}
	\begin{tabular}{lcccc}
	\toprule
	Dataset & Metric & \ourModel  -$\lambda$ & \ourModel-$\mathbf{Q}$ & \ourModel-$\mathcal{L}_{reg}$ \\
	\midrule
	\multirow{3}{*}{Cora} & 
	Accuracy &
	$\bm{-0.017}^*$ &
	$-0.007^*$ &
	$-0.012^*$ \\
	& Micro-F1 &
	$\bm{-0.021^*}$ &
	$-0.016^*$ &
	$-0.020^*$ \\
	& Macro-F1 &
	$\bm{-0.012}^*$ &
	$-0.007^*$ &
	$-0.008^*$ \\
	\midrule
	\multirow{3}{*}{Citeseer} & 
	Accuracy &
	$\bm{-0.004}$ &
	$-0.003$ &
	$\bm{-0.004}$ \\
	& Micro-F1 &
	$-0.008^*$ &
	$-0.003$ &
	$\bm{-0.009}^*$ \\
	& Macro-F1 &
	$-0.005$ &
	$-0.002$ &
	$\bm{-0.007}^*$ \\
	\midrule
	\multirow{3}{*}{Pubmed} & 
	Accuracy &
	$\bm{-0.017}^*$ &
	$-0.013^*$ &
	$-0.007^*$ \\
	& Micro-F1 &
	$\bm{-0.012}^*$ &
	$-0.009^*$ &
	$-0.003$ \\
	& Macro-F1 &
	$\bm{-0.012}^*$ &
	$-0.008^*$ &
	$-0.004$ \\
	\midrule
	\multirow{3}{*}{Facebook} & 
	Accuracy &
	$\bm{-0.020}^*$ &
	$-0.010^*$ &
	$-0.013^*$ \\
	& Micro-F1 &
	$\bm{-0.016}^*$ &
	$-0.011^*$ &
	$-0.012^*$ \\
	& Macro-F1 &
	$-0.012^*$ &
	$-0.009^*$ &
	$\bm{-0.015}^*$ \\
	\midrule
	\multirow{3}{*}{Youtube} & 
	Accuracy &
	$\bm{-0.007}^*$ &
	$-0.006$ &
	$-0.003$ \\
	& Micro-F1 &
	$\bm{-0.007}^*$ &
	$-0.005$ &
	$-0.002$ \\
	& Macro-F1 &
	$\bm{-0.009}^*$ &
	$-0.006$ &
	$-0.003$ \\
	\midrule
	\multirow{3}{*}{Amazon} & 
	Accuracy &
	$-0.008^*$ &
	$-0.003$ &
	$\bm{-0.012}^*$ \\
	& Micro-F1 &
	$-0.010^*$ &
	$-0.002$ &
	$\bm{-0.013}^*$ \\
	& Macro-F1 &
	$-0.007^*$ &
	$-0.003$ &
	$\bm{-0.009}^*$ \\
	\bottomrule
	\end{tabular}
	* suggests p-value < 0.05.
\end{table}

\subsection{Model Analysis}
\label{sec:model-analysis}
Finally, we provide an overall analysis on \ourModel, including an ablation study verifying the effectiveness of each component of it and a sensitivity study about the hyper-parameters related to membership modeling.

\noindent\textbf{$\bullet$ Ablation Study.}
We construct three variants of \ourModel\ by disabling one component at a time: 
1) \textbf{GraphHAM-$\lambda$} removes the membership-based attention coefficient $\lambda$ in the aggregation layer defined in Eq \eqref{eq:attention}; 
2) \textbf{GraphHAM-$\mathbf{Q}$} omits the group indicator $z_i$ in Eq \eqref{eq:loss-context} and replaces $\mathbf{Q}$ with a single vector $\mathbf{q}$, thus only preserves a membership-agnostic context;
3) \textbf{GraphHAM-$\mathcal{L}_{reg}$} removes the inter-layer membership regularization defined by Eq \eqref{eq:loss-reg}, and thus no inclusive constraint is imposed on groups across layers in it.
The performance of these variants compared with the complete model on node classification is summarized in Table \ref{tab:ablation}. 
We use Student's t-test to quantify the difference between the cross-validation results from the complete \ourModel{} model and each variant. The values marked with asteroid in the table suggest the difference is significant (i.e., p-value<0.05).
We can verify the importance of each component based on the gap in performance.
First, GraphHAM-$\lambda$ gave the worst results and the majority of performance values were significantly different from the complete model (marked by star), which highlights the importance of the membership-level attention in encoding the global information.
The second most effective design is the inter-layer regularization that explicitly forces an inclusive relation across layers to form a well-defined hierarchy.
The membership-dependent context decoder also improves the embedding quality, which suggests that even the same context perceived by nodes with different memberships reveal different information about node neighborhood. 

\begin{figure} 
\centering
\includegraphics[width=3.5in, height=2.8cm]{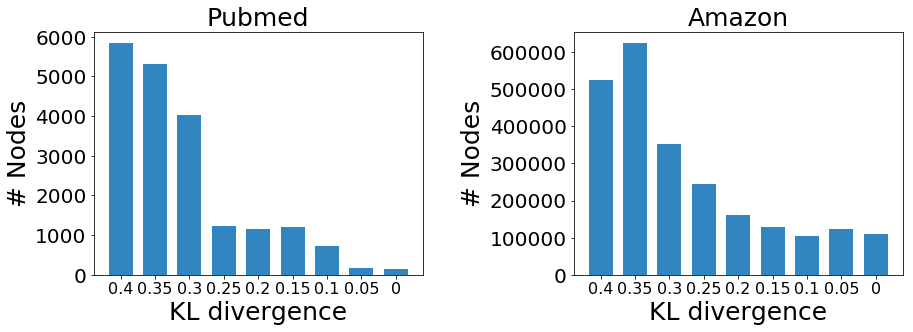}
\caption{Number of nodes fall into each interval of KL divergence between self attention coefficients $\lambda$ and group attention coefficients $\alpha$.}
\label{fig:kl} 
\end{figure}

\noindent\textbf{$\bullet$ Group attention versus node attention.}
Since the membership-based attention gives the most improvement, an interesting question to study is \textit{how different the group-level and the node-level attentions are}.
To quantify the information difference brought by these two types of attention, on each node $v_i$, we calculated the average KL-divergence between $\alpha$ and $\lambda$ as follows: $$\text{diff}(v_i)=\frac{1}{|\mathcal{N}_i|}\sum\nolimits_{j\in\mathcal{N}_i}KL(\alpha_{ij},\lambda_{ij})$$
We then count the number of nodes in different ranges of KL-divergence shown in Figure \ref{fig:kl}, where each bar $x$ collects nodes with $\text{diff}(v)\in[x,x+0.05)$.
It is clear that a large amount of nodes exhibit high degree of difference, which demonstrates that these two types of attention capture different aspects of node relatedness and are thus complementary to each other. 

\begin{figure} 
\centering
\includegraphics[width=3.5in, height=3.0cm]{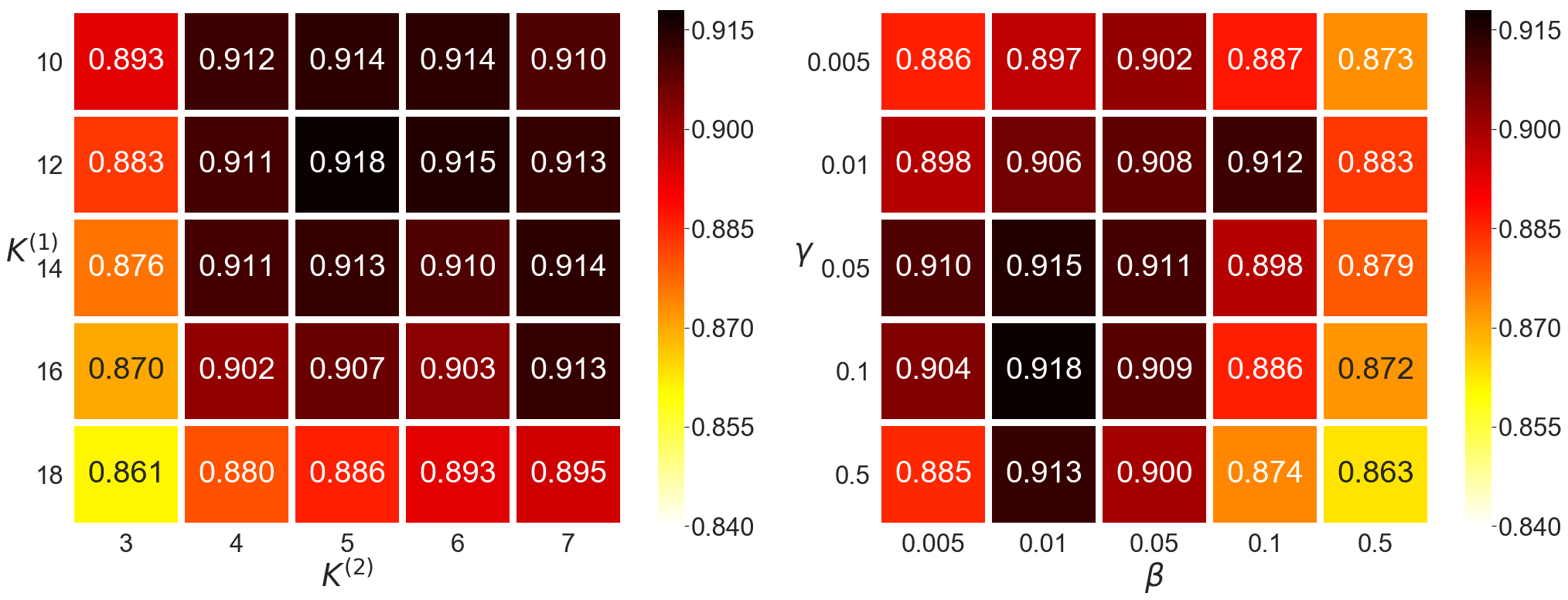}
\caption{Accuracy of node classification on Facebook graph under different hyper-parameter settings for group size $K^{(1)}, K^{(2)}$ (left) and regularization coefficients $\gamma, \beta$.}
\label{fig:parameter} 
\vspace{-2mm}
\end{figure}

\noindent\textbf{$\bullet$ Hyper-Parameter Sensitivity.}
We analyzed two groups of hyper-parameters related to membership modeling: the numbers of groups $K^{(1)}$ and $K^{(2)}$ for the two layers, and the weights of penalty $\gamma$ and $\beta$ for must- and cannot-link constraints.
Figure \ref{fig:parameter} reports the performance of \ourModel\ on Facebook graph under different hyper-parameter settings.
The model is generally more stable with respect to the number of groups, compared with the coefficients for the regularization term. Setting those hyper-parameters either too large or too small will compromise the performance.
More specifically, performance peaks when $K^{(1)}=12$ while $K^{(2)}=5$ on the Facebook graph, which has $4$ classes. Therefore, we believe a guidance for setting $K^{(2)}$ is to make it comparable with the number of classes while setting $K^{(1)}$ to be mildly larger than $K^{(2)}$. To balance the effect of the must- and cannot-link constraints, $\gamma$ should be larger than $\beta$ since the must-link set is usually much smaller.

%% file: conclusion.tex
\vspace{-0.1cm}
\section{Conclusion and Future Work}
In this paper, we proposed a new graph embedding model that captures a hierarchy of latent node groups and the nodes' affiliation to such groups.
We aligned the group hierarchy with the convolutional layers in GCNs, under the principle that lower layer aggregates neighborhood within a smaller scope thus captures fine-grained groups; and once the groups are merged at a higher level, more globally shared patterns and contexts are captured.
We designed a joint node-level and group-level attention mechanism, which brings both local and global patterns to aggregate the neighbor structure more accurately.
We trained the node embeddings by preserving a membership-based context, with inter-layer regularizations to inject inclusive relation among memberships.

As our future exploration, we are especially interested in the use of the membership vectors, which are shown to be a good measure for discovering nodes at the boundary of latent groups. This insight could broaden the impact and applicability of our model to serve more real-world problems.
For example, we can apply the membership profiles of nodes to detect the nodes at the boundary of ``changes'' to capture temporal patterns in dynamic graphs, or the nodes at the boundary of ``attacks'' to identify potential security issues in adversarial setting of graph embedding.